\documentclass[sigconf]{acmart}
\renewcommand\footnotetextcopyrightpermission[1]{}
\AtBeginDocument{%
  }

\setcopyright{acmlicensed}
\copyrightyear{2026}
\acmYear{2026}
\acmDOI{XXXXXXX.XXXXXXX}
\acmConference[ACM 2026]{Make sure to enter the correct conference title from your rights confirmation email}{Feb 1, 2026}{USA}
\acmISBN{978-1-4503-XXXX-X/2018/06}

\usepackage{multirow}
\usepackage{fancyvrb}

\newcommand{\squishlist}{
\begin{list}{$\bullet$}
  { \setlength{\itemsep}{0pt}
     \setlength{\parsep}{0pt}
     \setlength{\topsep}{0pt}
     \setlength{\partopsep}{0pt}
     \setlength{\leftmargin}{0em}
     \setlength{\labelwidth}{0em}
     \setlength{\labelsep}{0.2em} } }

\newcommand{\squishlisttwo}{
\begin{list}{$\bullet$}
  { \setlength{\itemsep}{0pt}
     \setlength{\parsep}{0pt}
    \setlength{\topsep}{0pt}
    \setlength{\partopsep}{0pt}
    \setlength{\leftmargin}{2em}
    \setlength{\labelwidth}{1.5em}
    \setlength{\labelsep}{0.5em} } }

\newcommand{\squishend}{
  \end{list}  }

\begin{document}


\title{Hard Rules, Soft Preferences: Bridging Reasoning, Learning, and Optimization for Personalized Packing Checklist Generation}

\author{Himel Dev\textsuperscript{1}, Madhusudan Basak\textsuperscript{2}, Tanmoy Sen\textsuperscript{3}, Paromita Shome\textsuperscript{2}, Bashima Islam\textsuperscript{2}}
\affiliation{%
  \institution{%
    \textsuperscript{1}529 Tech LLC,  \textsuperscript{2}University of Massachusetts Amherst, \textsuperscript{3}University of Virginia \\
    himel@529-tech.com, mbasak@umass.edu, ts5xm@virginia.edu, pshome@umass.edu, bashima@umass.edu
  }
  \country{USA}
}

\begin{abstract}
Packing for air travel is recurring and error-prone: the checklist must be personal and context-aware, yet feasible under safety rules, item dependencies, and luggage limits. Existing packing assistants are template-driven (generic) or recommendation-driven but unconstrained, leaving users to manually patch regulatory and capacity violations. We propose a reasoning-guided learning framework with three stages: (1) a symbolic engine that generates a regulation-aware seed checklist with explicit dependency structure, (2) a two-stage preference learner that estimates inclusion and priority utilities from user add and remove actions, mitigating survivorship bias, and (3) a CP-SAT optimizer that selects a compact, compliant subset. The architecture instantiates a general pattern for constrained personalization, applicable wherever hard feasibility coexists with sparse preference signals. On 604 labeled trip scenarios (29K inclusion labels, 343K pairwise comparisons), the symbolic engine attains 99.7\% recall and 0.96 rubric validity versus 0.78 to 0.81 for frontier LLMs; gradient boosted trees and LambdaMART reach AUC-ROC 0.943 and NDCG@5 0.923; and CP-SAT attains 100\% constraint satisfaction versus 28\% for greedy and 10\% for random. Deployment in FlyEnJoy, a production iOS travel app, doubled checklist completions and reduced editing and completion time. To support reproducibility, we publicly release a Python implementation of the checklist system along with sample data.\footnote{\url{https://github.com/Official529Tech/rlo-checklist}}
\end{abstract}


\maketitle

\section{Introduction}
Packing for air travel is a recurring, error-prone task with non-negotiable regulatory~\cite{tsa2026whatcanibring}, safety~\cite{tsa2026advisories}, capacity, and dependency constraints. For each trip, travelers must adapt to the travel context (destination, duration, activities, special needs) while respecting these constraints alongside personal preferences. Missing essentials, overpacking, or carry-on violations are common and lead to inconvenience, added cost, and compliance or safety risks.

\begin{figure}[t]
    \centering
    \includegraphics[width=0.8\linewidth]{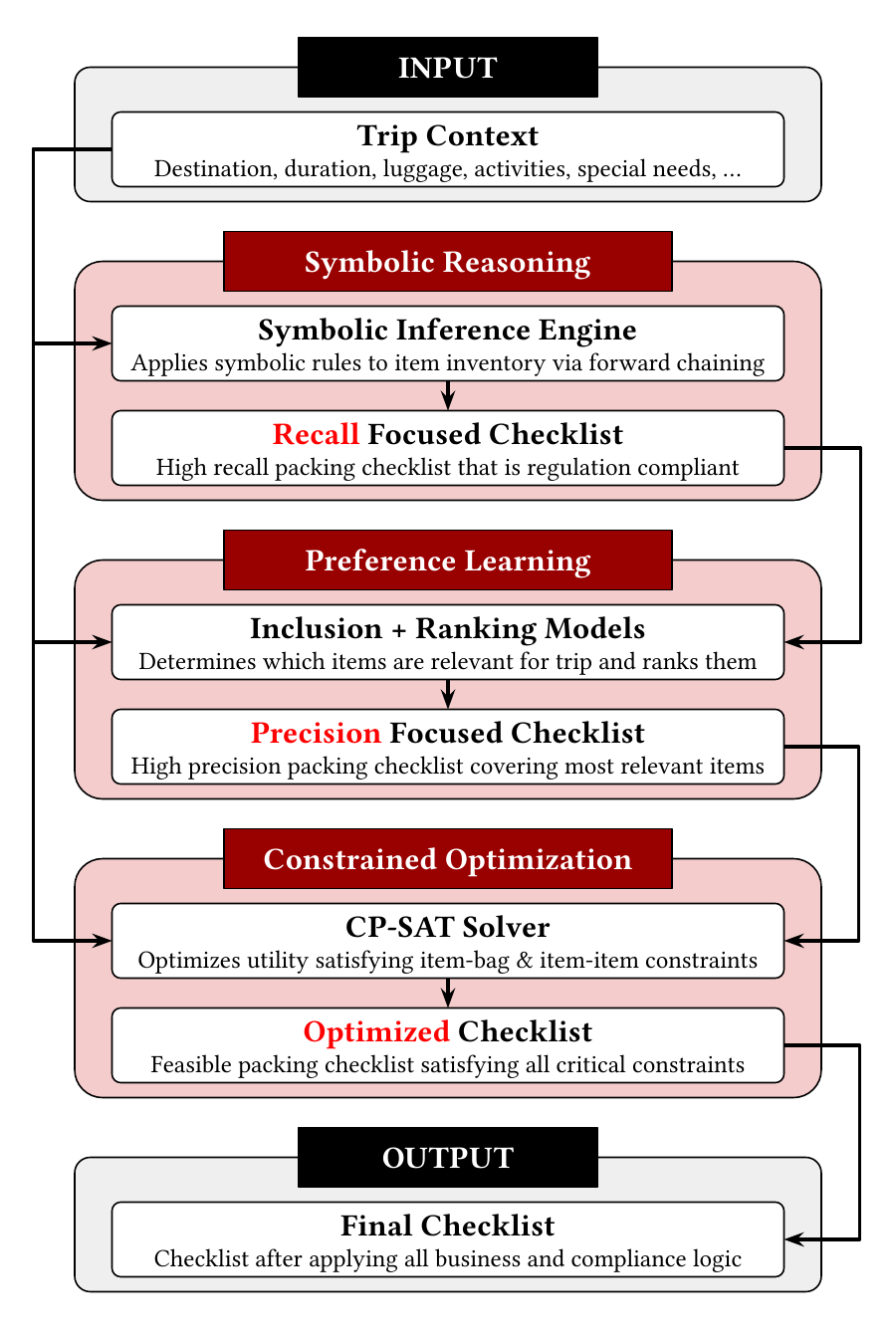}
    \vspace{-1em}
    \caption{System architecture. Stage 1 generates a high-recall, regulation-compliant seed checklist via symbolic rules. Stage 2 learns relevance, priority, and context-aware utilities from user edits. Stage 3 selects and assigns items under hard constraints via constrained optimization.}
    \label{fig:system_overview}
    \vspace{-1.5em}
\end{figure}

A 2024 survey of 2{,}000 U.S. adults found that nine in ten respondents realized en route that they had forgotten something essential, most often phone chargers, toiletries, sunscreen, or medication~\cite{talkerresearch2024forget}. Forty-two percent reported that these omissions negatively impacted their travel experience, underscoring the need for decision support beyond memory or static lists. Carry-on liquid limits and battery-related restrictions~\cite{tsa2026whatcanibring,faa2026lithium} further raise the bar: a useful system must guarantee feasibility, not merely suggest items.

For travelers with special needs, compliance failures happen at the security checkpoint, not merely at the destination. A parent traveling with an infant can carry breast milk and formula above standard carry-on liquid limits, but only if they declare them for separate TSA screening. An insulin traveler must keep accompanying ice packs frozen solid at the checkpoint, or they fall under the 3-1-1 rule. A pregnant traveler with liquid medication must invoke the TSA medical exemption and declaration procedure. These are not preferences but regulatory requirements that vary by traveler profile and cannot be captured by one-size-fits-all templates.

Despite this need, existing packing assistants~\cite{lim2024enhancing,nikam2025interactive,ravi2025multi,regin2024swifttrip} largely fail to model feasibility and offer only limited personalization. Template-based checklists~\cite{christine2024ultimate,Canva2026free} are easy to deploy but ignore context (e.g., thermal wear for cold climates), personal preferences, and inter-item dependencies (e.g., a camera implying chargers and memory cards). Learning-based recommenders can personalize but are hampered by sparse per-user histories, wide trip variability, and largely implicit feedback. Critically, unconstrained recommenders cannot guarantee feasibility under the hard constraints introduced above, and naive post-filtering breaks coverage or dependencies when constraints interact. State-of-the-art large language models (LLMs) fare no better. In our evaluation (Section~\ref{sec:evaluation}), frontier LLMs achieve only 78 to 81 percent validity on a rubric covering TSA compliance, weather appropriateness, activity needs, and completeness, compared to 96 percent for our symbolic engine.

To address this, we propose a reasoning-guided learning framework that generates personalized packing checklists while guaranteeing feasibility under hard constraints. The system (Figure \ref{fig:system_overview}) decomposes the task into three stages, each addressing a distinct concern that template-only and unconstrained recommendation approaches handle poorly: (1) \textbf{Symbolic Reasoning} — a symbolic engine encoding 226 domain rules over 378 items to produce a high-coverage seed checklist consistent with safety and regulatory requirements and robust under cold-start; (2) \textbf{Preference Learning} — a two-stage model that learns trip-specific utility from user-edit logs, separating inclusion from priority ordering to mitigate survivorship bias; and (3) \textbf{Constrained Optimization} — a CP-SAT optimizer that selects and packs items to maximize learned utility subject to capacity, safety, regulatory, environmental, and dependency constraints.


We deploy and evaluate the system within FlyEnJoy,\footnote{\url{https://apps.apple.com/app/6745104853}} a production iOS travel app that runs entirely on-device to support travelers with limited connectivity. The on-device constraint rules out frontier LLM serving and motivates the lightweight, interpretable models used in Stage~2. FlyEnJoy provides itinerary generation, packing checklists, and document management; the packing module instantiates the framework described in this paper.

We make the following contributions:
\squishlist
    \item \textbf{Constraint-Aware Candidate Generation.} A symbolic reasoning engine that achieves 99.7\% candidate recall and outperforms frontier LLMs on rubric-based validity (0.96 vs.\ 0.78 to 0.81), with rule provenance enabling cold-start generalization to rare items.

    \item \textbf{Survivorship-Bias-Aware Preference Learning.} A two-stage formulation that separates item inclusion from priority ranking to mitigate survivorship bias in implicit checklist feedback, achieving AUC-ROC 0.943 for inclusion and NDCG@5 0.923 for ranking.

    \item \textbf{By-Construction Feasibility via Constrained Optimization.} A CP-SAT formulation for constrained checklist selection and bag assignment that guarantees 100\% feasibility under interacting compliance, capacity, and dependency constraints, compared to 28\% for greedy and 10\% for random baselines.
\squishend

Deployment in FlyEnJoy doubled checklist completions and reduced editing and completion time, confirming real-world utility. While our instantiation addresses air travel packing, the three-stage architecture is a general design pattern for constrained personalization, applicable wherever hard feasibility requirements coexist with sparse preference signals. Other instances include clinical discharge planning and immigration documentation.

\section{Problem Formulation}
\label{sec:problem}
We formulate personalized packing checklist generation as a constrained subset-selection problem under contextual, preference, and feasibility considerations.

\noindent\textbf{Inputs.}
Each trip instance is characterized by (i) trip context $c$ (origin, destination, travel dates, purpose, accommodation, luggage, activities, and special needs), and (ii) a set of hard constraints $\mathcal{R}$ capturing item eligibility (regulatory and safety rules), physical luggage capacity, and inter-item dependencies.

\noindent\textbf{Output.}
The system outputs a checklist $S \subseteq \mathcal{I}$ over a universe of items $\mathcal{I}$, optionally with a bag assignment $b: S \rightarrow \mathcal{B}$. The triple $(c, \mathcal{R}, S)$ instantiates personalized constrained subset-selection task.

\noindent\textbf{Objective.}
The goal is to generate a checklist that maximizes predicted traveler utility, $\max_{S} \; U(S \mid c) \;\text{s.t.}\; S \text{ is feasible under } \mathcal{R}$. Here, $U(S \mid c)$ aggregates item-level utilities learned in Section~\ref{sec:preference}.

\noindent\textbf{Desiderata.}
A practical packing system must satisfy:
\squishlist
    \item \textbf{Compliance:} All regulatory (e.g., TSA) and safety (e.g., medical) requirements must be met without exception.
    \item \textbf{Recall:} Essential items (e.g., phone charger) must be included even when user information is sparse.
    \item \textbf{Personalization:} Recommendations should adapt to trip-specific factors such as destination, weather, and activities.
    \item \textbf{Precision:} The checklist should avoid unnecessary or redundant items (e.g., a heavy coat for a tropical destination).
    \item \textbf{Feasibility:} Physical capacity, bag eligibility, and inter-item dependency constraints must be respected.
    \item \textbf{Utility:} Packing decisions should balance expected retained utility against risk of bag loss.

\squishend


\section{System Overview}
\label{sec:system}
The system consists of three stages, each addressing a distinct concern and feeding the next.

\noindent\textbf{Stage 1: Symbolic Reasoning (Compliance and Recall).}
Given a trip context, a symbolic inference engine applies a prioritized set of declarative rules that encode regulatory requirements, safety constraints, contextual triggers, and item dependencies. The output is a high-recall, regulation-aware seed checklist that remains comprehensive even under cold-start.


\noindent\textbf{Stage 2: Preference Learning (Personalization and Precision).}
Given the seed checklist, a two-stage learning pipeline estimates (i) per-item inclusion probabilities and (ii) relative priority among retained items, separating the two signals to mitigate survivorship bias. Features combine trip context with symbolic provenance (i.e., the rules and triggers that proposed each item), enabling robust preference learning under sparse interaction data.

\noindent\textbf{Stage 3: Constrained Optimization (Feasibility and Utility).}
The final stage formulates item selection and bag assignment as a constrained optimization problem. A CP-SAT solver selects a utility-maximizing subset of items and assigns them to bags subject to regulatory, capacity, and dependency constraints, with risk of bag loss reflected in the objective.


\subsection{Symbolic Reasoning}

The first stage employs a symbolic reasoning engine to generate a high-recall, regulation-compliant seed checklist. We prioritize compliance, recall, and explainability over personalization at this stage because purely data-driven recommendations are unreliable under cold-start, where infrequent travel and limited user history yield sparse, biased signals. The engine encodes 226 declarative rules spanning regulatory, safety, contextual, and traveler-specific knowledge, operating over catalogs of 378 atomic items and 103 pre-trip tasks. Regulatory rules are sourced from authoritative TSA and FAA guidance~\cite{tsa2026whatcanibring,faa2026lithium}.


\noindent\textbf{Knowledge Representation.}
Domain knowledge is encoded as a collection of declarative, human-readable rules authored by internal specialists. Each rule specifies a \emph{when} clause over enriched trip context and inferred facts, and a \emph{then} clause that emits recommended items, pre-trip tasks, or additional facts. Rules are organized into seven categories covering regulatory concerns, wellbeing, essentials, contextual triggers, activities, traveler-specific factors, and inter-item dependencies (Table~\ref{tab:rules}).
Each rule carries a priority reflecting its criticality: regulatory and safety rules (e.g., travel documents, liquids, batteries) are evaluated before contextual and comfort-oriented rules. This priority structure makes inference deterministic and auditable: safety- and legality-critical inferences form the factual foundation on which later rules depend.

\begin{table}[t]
\centering
\small
\caption{Structure and priority distribution of symbolic rules. Higher-priority rules enforce regulatory and safety constraints before contextual and traveler-specific reasoning.}
\vspace{-0.1in}
\resizebox{\columnwidth}{!}{
\begin{tabular}{lccll}
\toprule
\textbf{Rule Category} & \textbf{\# Rules} & \textbf{Priority} & \textbf{Primary Triggers} & \textbf{Examples} \\
\midrule
Regulatory   & 52  & High   & Destination, Luggage  & TSA/FAA regulations \\
Wellbeing    & 38  & High   & Special needs         & Medical needs \\
Essential    & 24  & Medium & Trip duration         & Basic items \\
Contextual   & 39  & Medium & Destination, Lodging  & Weather adaptation \\
Activities   & 31  & Medium & Activities            & Attending events \\
Dependencies & 7   & Medium & Item-derived facts    & Complementary items \\
Traveler     & 35  & Low    & Age, Gender           & Gender-specific needs \\
\midrule
\textbf{Total} & \textbf{226} & -- & -- & -- \\
\bottomrule
\end{tabular}
}
\label{tab:rules}
\end{table}

\noindent\textbf{Context Enrichment.}
Raw user inputs (e.g., origin, destination, start date, duration, purpose) are normalized and enriched into a structured context. The system automatically derives higher-level attributes such as travel type (domestic vs.\ international) from the origin-destination pair and temperature category from weather data specific to destination and date. Multi-select inputs such as activities and special needs are converted into atomic facts (e.g., \texttt{special\_need(Infant)}, \texttt{has\_activity(Hiking)}), enabling simultaneous activation of multiple rule categories.
During inference, rules may also emit new facts (e.g., \texttt{has\_liquids}, \texttt{requires\_adapter}), enabling cascading inferences in which earlier rule outputs establish preconditions for subsequent regulatory or dependency rules.

\noindent\textbf{Forward-Chaining Inference.}
Inference uses priority-based forward chaining: starting from facts derived from the enriched context, rules fire when their conditions are satisfied, emitting additional facts, item candidates, and pre-trip tasks until a fixpoint is reached. In practice, convergence occurs within a small number of iterations bounded by the rule dependency graph. The procedure supports cascading inferences (e.g., toiletries $\rightarrow$ liquids $\rightarrow$ TSA quart-size bag) and is deterministic: identical inputs yield identical outputs.
Multiple rules may recommend the same item under different justifications or criticality levels. We merge such duplicates into a single candidate by conservative criticality escalation (mandatory $>$ recommended $>$ optional), least-restrictive feasible placement, and union of triggering reasons. The resulting provenance record (rules fired, priorities, triggering conditions) provides structured attribution signals used as features in Stage~2.


\subsection{Preference Learning}
\label{sec:preference}
The second stage learns \emph{soft} personalization signals over the rule-generated seed checklist. While the symbolic stage deliberately over-generates a high-recall, regulation-compliant seed, practical packing demands \emph{parsimony}: the second stage selects relevant items and prioritizes them in trip context, producing item-level utilities for downstream optimization.
We model preferences as two coupled subproblems: (i) inclusion -- whether a suggested item should be packed at all -- and (ii) ranking -- relative priority among included items. Decomposing into inclusion and ranking is essential to avoid \emph{survivorship bias}: ranking labels are observed only for items the labeler retained, so a single ranker would systematically overestimate the utility of items frequently removed by other labelers.

\noindent\textbf{Data Collection.}
\label{sec:data_collection}
We collected preference annotations with our web-based packing evaluation platform that mirrors the interface and interaction flow of our production application, enabling controlled, repeatable annotation of naturalistic packing decisions.

\noindent\textit{Annotation Procedure.}
For each annotation task $t$, a labeler is presented with a trip context $c_t$ (the inputs defined in Section~\ref{sec:problem}). The symbolic engine generates a high-recall seed checklist $S_t \subset \mathcal{I}$. Labelers then remove irrelevant items, add missing ones, and rank the remainder by perceived importance, producing an edited checklist $F_t \subset \mathcal{I}$ with associated total order $\pi_t$.

\noindent\textit{Derived Signals.}
From each annotated trip, we derive three disjoint item sets: (i) kept items, $K_t = S_t \cap F_t$, (ii) removed items, $D_t = S_t \setminus F_t$, (iii) added items, $A_t = F_t \setminus S_t$. Across annotated trip scenarios, this process yields implicit preference supervision conditioned on trip context, without requiring explicit ratings or repeated labeling of the same scenario.

\noindent\textbf{Two-Stage Modeling.}
\label{sec:two_stage_signals}
Labeler edits provide two complementary but distinct preference signals that must be modeled separately to avoid survivorship bias:

\noindent\textit{Signal 1: Binary Inclusion Decisions.}
For each item $i \in S_t$, the labeler's decision to keep or remove the item reveals whether it is contextually relevant and aligned with their packing style. This provides binary supervision: $y^{\mathrm{include}}_{t,i} = \mathbb{1}[i \in F_t]$.

\noindent\textit{Signal 2: Relative Ranking among Kept Items.}
Among items retained in $F_t$, the complete ranking $\pi_t$ provides fine-grained priorities. For any ordered pair $(i,j)$ where $i$ is ranked above $j$ in $\pi_t$, we record relative importance $i \succ_t j$.

\noindent\textit{Why Separation Matters: Survivorship Bias.}
Ranking labels are only observed for items that the labeler retained. If we trained a ranker alone, items frequently removed by many labelers would be absent from pairwise comparisons, inflating their apparent utility when they do appear. For example, a travel pillow may rank mid-to-high among retained items for comprehensive packers, yet be removed entirely by minimalists; modeling only rankings would overestimate its unconditional utility. Decomposing personalization into inclusion (``pack or not'') and conditional ranking (``priority if packed'') prevents this survivorship bias and yields utilities that are meaningful for downstream selection under capacity constraints. 

\noindent\textbf{Inclusion Model (Precision Layer).}
\label{sec:inclusion_model}
The inclusion model predicts the probability that a rule-suggested item is retained after labeler editing, conditioned on trip context and symbolic provenance.

\noindent\textit{Model Architecture.}
For each item $i \in S_t$, we model:
\begin{equation}
p^{\mathrm{include}}(i \mid c_t, \phi_{t,i})
=
\sigma\!\left(F_M\!\left(\psi^{\mathrm{inc}}(i, c_t, \phi_{t,i})\right)\right),
\end{equation}
where $\sigma$ is the sigmoid function and $F_M$ is an additive ensemble of $M$ regression trees. We use gradient boosted trees (GBM), which capture non-linear feature interactions and provide interpretability through feature importance analysis.

\noindent\textit{Feature Representation.}
We represent each candidate by (i) symbolic provenance features (e.g., rule family, max rule priority, criticality, and safety/regulatory flags), (ii) trip context features (e.g., duration bin, destination/temperature type, activities, luggage type), and (iii) item-context interaction features. 

\noindent\textit{Training Objective.}
Given trip data $\mathcal{D} = \{(S_t, F_t, c_t)\}_{t=1}^N$, we minimize binary cross-entropy (BCE) via gradient boosting:
\begin{equation}
\mathcal{L}_{\mathrm{include}}
=
\sum_{t=1}^N \sum_{i\in S_t}
\mathrm{BCE}\!\left(y^{\mathrm{include}}_{t,i},\;
p^{\mathrm{include}}(i \mid c_t,\phi_{t,i})\right),
\end{equation}
regularized through tree depth ($\mathrm{depth}=5$) and shrinkage (learning rate $\eta=0.1$). Rule-family indicators in $\phi_{t,i}$ allow the inclusion model to learn shared acceptance patterns across items contributed by the same rule family, improving robustness for rare items.

\noindent\textbf{Ranking Model (Priority Layer).}
\label{sec:ranking_model}
Among items predicted to be kept, the ranking model learns their relative importance and priority ordering.

\noindent\textit{Model Architecture.}
We employ a learning-to-rank approach using LambdaMART \cite{burges2010ranknet}, implemented via LightGBM's ranking objective. For any pair of items $(i,j)$ both appearing in final list $F_t$, we model:
\begin{equation}
p^{\mathrm{rank}}_{\eta}(i \succ_t j \mid c_t, \phi_{t,i}, \phi_{t,j})
=
\sigma\!\left(s_{\eta}(i, c_t, \phi_{t,i}) - s_{\eta}(j, c_t, \phi_{t,j})\right),
\end{equation}
where $s_{\eta}(i, c_t, \phi_{t,i})$ is a gradient-boosted scoring function that assigns a real-valued relevance score to item $i$ in context $c_t$.

\noindent\textit{Pairwise Feature Representation.}
For each ordered pair $(i,j)$ with $i \succ_t j$, we compile pairwise difference features (e.g., priority and criticality deltas), categorical comparison features, and absolute provenance and contextual features. 

The ranker learns to combine these signals to predict relative preferences. Difference features capture symmetric comparisons, while absolute features allow position-dependent effects (e.g., mandatory items tend to rank high regardless of the comparison).

\noindent\textit{Training Objective.}
Let $\mathcal{P}_t = \{(i,j) : i,j \in F_t, \, i \succ_t j\}$ denote the set of pairwise preferences from trip/task $t$. We minimize LambdaRank loss \cite{burges2010ranknet}:
\begin{equation}
\mathcal{L}_{\mathrm{rank}}(\eta)
=
\sum_{t=1}^N \sum_{(i,j) \in \mathcal{P}_t}
\log\!\left(1 + \exp\left(-(s_{\eta}(i,c_t,\phi_{t,i}) - s_{\eta}(j,c_t,\phi_{t,j}))\right)\right),
\end{equation}
weighted by the change in Normalized Discounted Cumulative Gain (NDCG) induced by swapping $i$ and $j$ in the ranking. This loss prioritizes correcting mistakes at the top of the ranking, where user attention is concentrated.

We use gradient boosting with shallow trees (max depth = 4) and L2 leaf regularization to prevent overfitting.

\noindent\textbf{Combined Utility for Optimization.}
\label{sec:combined_utility}
The learned inclusion probabilities and ranking scores are combined into a unified utility function for downstream constraint-aware optimization.

\noindent\textit{Multiplicative Composition.}
For each item $i$ in seed checklist $S_t$, we define:
\begin{equation}
u_i(c_t)
=
p^{\mathrm{include}}_{\theta}(i \mid c_t, \phi_{t,i})
\cdot
s_{\eta}(i, c_t, \phi_{t,i})
\cdot
u^{\mathrm{sym}}_i(c_t),
\end{equation}
where $p^{\mathrm{include}}_{\theta}$ captures the probability that the item is contextually relevant, $s_{\eta}$ captures its relative priority among relevant items, and $u^{\mathrm{sym}}_i(c_t)$ is a symbolic prior enforcing safety and regulatory requirements.

\noindent\textit{Symbolic Prior.}
The symbolic utility $u^{\mathrm{sym}}_i(c_t)$ ensures that safety-critical and legally required items receive elevated utility regardless of learned preferences:
\begin{equation}
u^{\mathrm{sym}}_i(c_t) = 
\begin{cases}
10.0 & \text{if } i \text{ is regulatory-required (e.g., passport)} \\
5.0 & \text{if } i \text{ is safety-critical (e.g., medication)} \\
1.0 & \text{otherwise}
\end{cases}
\end{equation}

This multiplicative decomposition preserves hard guarantees from symbolic reasoning (e.g., regulatory items always receive high utility) while enabling fine-grained personalization driven by learned preferences (e.g., distinguishing between multiple optional items based on context).

\subsection{Constrained Optimization}
\label{sec:optimization}

We consider the problem of assigning a set of travel items to a limited number of heterogeneous bags in order to maximize retained utility, while respecting physical feasibility, regulatory requirements, environmental suitability, inter-item dependencies, and risk considerations.
The problem captures the tension between strict feasibility constraints (e.g., airline regulations and bag capacities) and softer, preference-driven factors (e.g., convenience, fragility, and loss risk).

\noindent\textbf{Problem Setting.}
Let $\mathcal{I}$ denote a set of items and $\mathcal{B}$ denote a set of bags (e.g., checked baggage, carry-on, personal item).
Each item $i \in \mathcal{I}$ is characterized by:
(i) physical dimensions $(l_i, w_i, h_i)$ and weight $\mathrm{wt}_i$,
(ii) a perceived utility $u_i \ge 0$,
(iii) a monetary value $v_i \ge 0$ (representing replacement cost or intrinsic worth), and
(iv) attributes governing regulatory eligibility, environmental sensitivity, and convenience.

Each bag $b \in \mathcal{B}$ is defined by:
(i) internal dimensions $(L_b, W_b, H_b)$,
(ii) a maximum allowable weight $C_b$, and
(iii) a loss probability $p_b \in [0,1]$, reflecting the relative risk of loss or damage (e.g., higher for checked baggage).

The objective is to select a subset of items and assign each selected item to at most one bag so as to maximize expected retained utility while minimizing expected monetary loss, subject to feasibility and preference constraints.

\noindent\textbf{Decision Variables.}
We introduce the following decision variables: (i) $x_{ib} \in \{0,1\}$, which equals $1$ if item $i$ is assigned to bag $b$; (ii) $y_i \in \{0,1\}$, which equals $1$ if item $i$ is packed in any bag. These variables are linked by $y_i = \sum_{b \in \mathcal{B}} x_{ib}$, where $\forall i \in \mathcal{I}$ ensuring that each item is assigned to at most one bag.

\noindent\textbf{Physical Feasibility Constraints.}
For each bag $b$, total weight cannot exceed capacity, $\sum_{i} \mathrm{wt}_i \, x_{ib} \le C_b$. Items must also fit dimensionally, which we model in two parts: (i) \emph{orientation compatibility}: item $i$ fits in bag $b$ only if its sorted dimensions are component-wise no larger than $b$'s, otherwise $x_{ib}=0$; and (ii) \emph{volume capacity}:
\begin{equation}
    \sum_{i} \mathrm{vol}_i \, x_{ib} \le \eta\, V_b, \quad \forall b \in \mathcal{B},
\end{equation}
where $\eta \in [0.7, 0.9]$ is a packing efficiency factor that controls realistic volume utilization.







\noindent\textbf{Regulatory and Environmental Constraints.}
Let $A^{\mathrm{reg}}_{ib} \in \{0,1\}$ indicate whether item $i$ is legally permitted in bag $b$ (e.g., spare lithium-ion batteries are barred from checked baggage), and $A^{\mathrm{env}}_{ib} \in \{0,1\}$ whether bag $b$ is environmentally suitable for $i$. We enforce:
\begin{equation}
    x_{ib} \le A^{\mathrm{reg}}_{ib}, \quad \forall i, b, \qquad
    x_{ib} \le A^{\mathrm{env}}_{ib}, \quad \forall i \in \mathcal{I}_{\mathrm{strict}}, b,
\end{equation}
where $\mathcal{I}_{\mathrm{strict}} \subseteq \mathcal{I}$ contains items with strict environmental needs (e.g., insulin requiring temperature control). Moderately sensitive items (e.g., chocolate, aerosol sunscreen) are not hard-constrained but penalized via $c^{\mathrm{env}}_{ib}$ in the objective.




\noindent\textbf{Inter-Item Dependency Constraints.}
Some item groups must be packed together (e.g., camera body, battery, lens, which are useless apart) or kept apart (e.g., liquids and electronics, to mitigate spillage). Let $\mathcal{G}_{\mathrm{together}}, \mathcal{G}_{\mathrm{apart}} \subseteq 2^{\mathcal{I}}$ be the corresponding collections of item groups. We enforce:
\[
    x_{ib} = x_{jb} \;\; \forall i,j \in G, \;\forall b,\; G \in \mathcal{G}_{\mathrm{together}}; \quad
    \sum_{i \in G} x_{ib} \le 1 \;\; \forall b,\; G \in \mathcal{G}_{\mathrm{apart}}.
\]


\noindent\textbf{Risk-Aware Objective.}
Each bag $b$ has loss probability $p_b \in [0,1]$. Two quantities drive the objective: expected retained utility (item is transported with probability $1-p_b$) and expected monetary loss (proportional to item value $v_i$ in lost bags):
\begin{equation}
    \underbrace{\sum_{i,b} u_i (1-p_b)\, x_{ib}}_{\text{retained utility}}, \qquad
    \underbrace{\sum_{b} p_b \sum_{i} v_i\, x_{ib}}_{\text{expected loss}}.
\end{equation}
This formulation discourages placing high-value items in high-risk bags (e.g., checked baggage), with the full trade-off combined in the overall objective below.




\noindent\textbf{Overall Optimization Problem.}
We formulate packing as a CP-SAT problem that enforces all preceding constraints as hard and the following preferences as soft penalties:
\begin{equation}
\begin{aligned}
\max \;\;
& \underbrace{
\sum_{i, b}
u_i (1 - p_b)\, x_{ib}
}_{\text{expected retained utility}}
- \underbrace{
\alpha \sum_{b} p_b \sum_{i} v_i \, x_{ib}
}_{\text{expected monetary loss}}
\\[4pt]
& - \underbrace{
\sum_{i \in \mathcal{I} \setminus \mathcal{I}_{\mathrm{strict}}, \, b}
c^{\mathrm{env}}_{ib}\, x_{ib}
}_{\text{environmental penalty}}
- \underbrace{
\sum_{i, b}
c^{\mathrm{soft}}_{ib}\, x_{ib}
}_{\text{convenience and fragility penalty}},
\end{aligned}
\end{equation}
where $c^{\mathrm{env}}_{ib} = c^{\mathrm{env}}_i \cdot (1 - A^{\mathrm{env}}_{ib})$ penalizes moderately sensitive items in unsuitable bags, $c^{\mathrm{soft}}_{ib}$ encodes bag-specific accessibility and fragility costs, and $\alpha \ge 0$ trades off expected loss against expected utility. CP-SAT~\cite{rossi2006handbook} natively handles the resulting mix of linear, logical, cardinality, equality, and bound constraints.

\section{Evaluation}
\label{sec:evaluation}
We evaluate the framework on labeled trip scenarios collected through a web-based annotation platform. The evaluation addresses four questions: (1) Does the symbolic engine recall the items labelers ultimately select? (2) How accurately do the preference models predict item inclusion and relative priority? (3) Does the CP-SAT optimizer satisfy all hard constraints where baselines fail? (4) How close does the end-to-end pipeline come to the oracle-feasible recall under the optimizer's capacity budget?

\subsection{Dataset and Experimental Setup}
We collected interaction data from labelers across 604 labeled trip scenarios. Each labeler was presented with a series of hypothetical trip contexts specifying origin, destination, duration, purpose, accommodation, luggage type, activities, and special needs. For each scenario, labelers received a seed checklist generated by the symbolic engine and were asked to curate a checklist as if they were packing for that trip. Labelers edited the checklist by removing unwanted items, optionally adding missing items, and providing a complete ranking over their final selections. Because labelers provide a \emph{complete} ranking over their final selections, 604 scenarios produce a substantial training signal: 29K binary inclusion labels and 343K pairwise ranking comparisons. 


Our item catalog contains 378 unique items governed by 226 symbolic rules. The symbolic engine generates seed checklists averaging $48.9 \pm 20.5$ items, which labelers reduce to $32.2 \pm 11.4$ items on average (acceptance rate $75.7\%$). Labelers added an average of only $0.08$ items beyond the seed, confirming the symbolic rules' near-complete coverage of relevant items.


\subsection{Stage 1: Symbolic Engine}
\begin{figure*}[!htb]
\begin{minipage}{0.31\textwidth}
\centering
\includegraphics[width=\columnwidth]{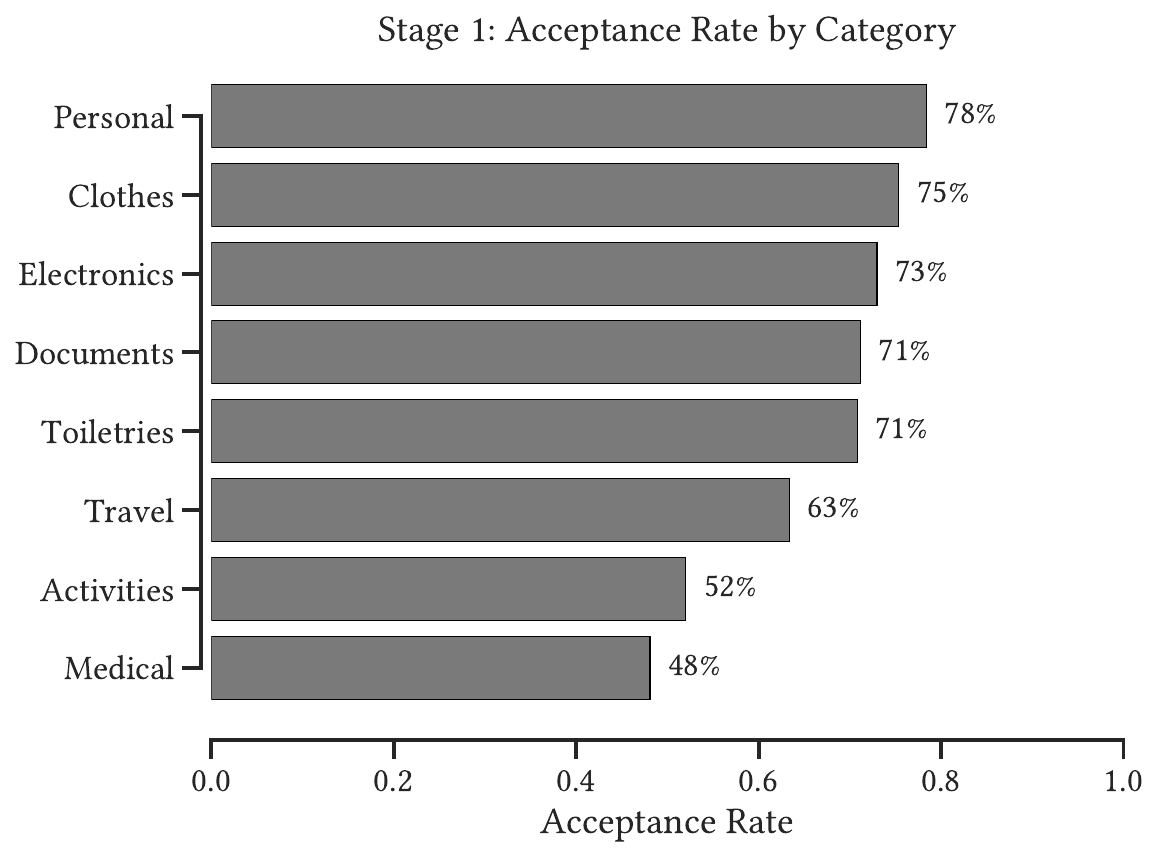}\vspace{-0.05in}
\caption{Acceptance rate by item category. Personal items have highest acceptance; medical items have lowest.\vspace{-0.1in}}
\label{fig:acceptance_category}
\end{minipage}
\hspace{.1em}
\begin{minipage}{0.31\textwidth}
\centering
\includegraphics[width=\columnwidth]{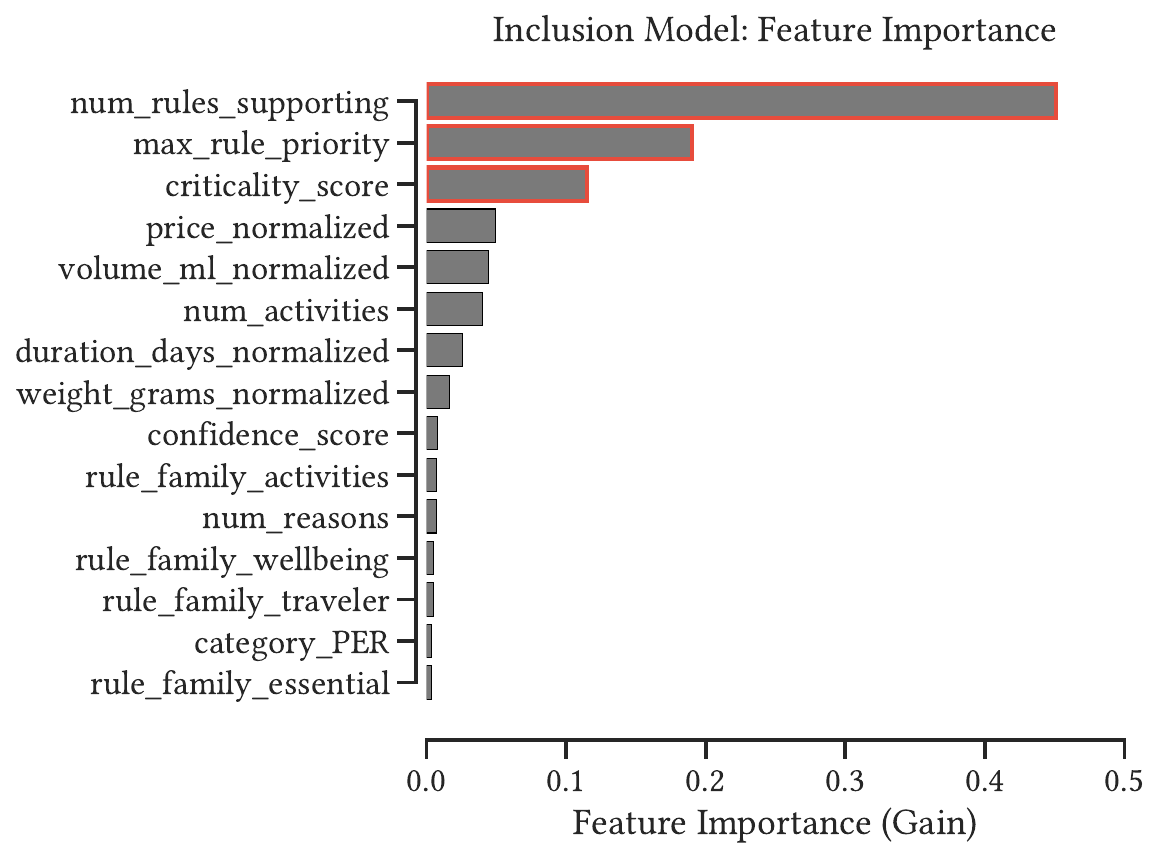}\vspace{-0.05in}
\caption{Inclusion model feature importance by gain. Symbolic provenance features are the strongest predictors.\vspace{-0.1in}}
\label{fig:inclusion_features}
\end{minipage}
\hspace{.1em}
\begin{minipage}{0.31\textwidth}
\centering
\includegraphics[width=\columnwidth]{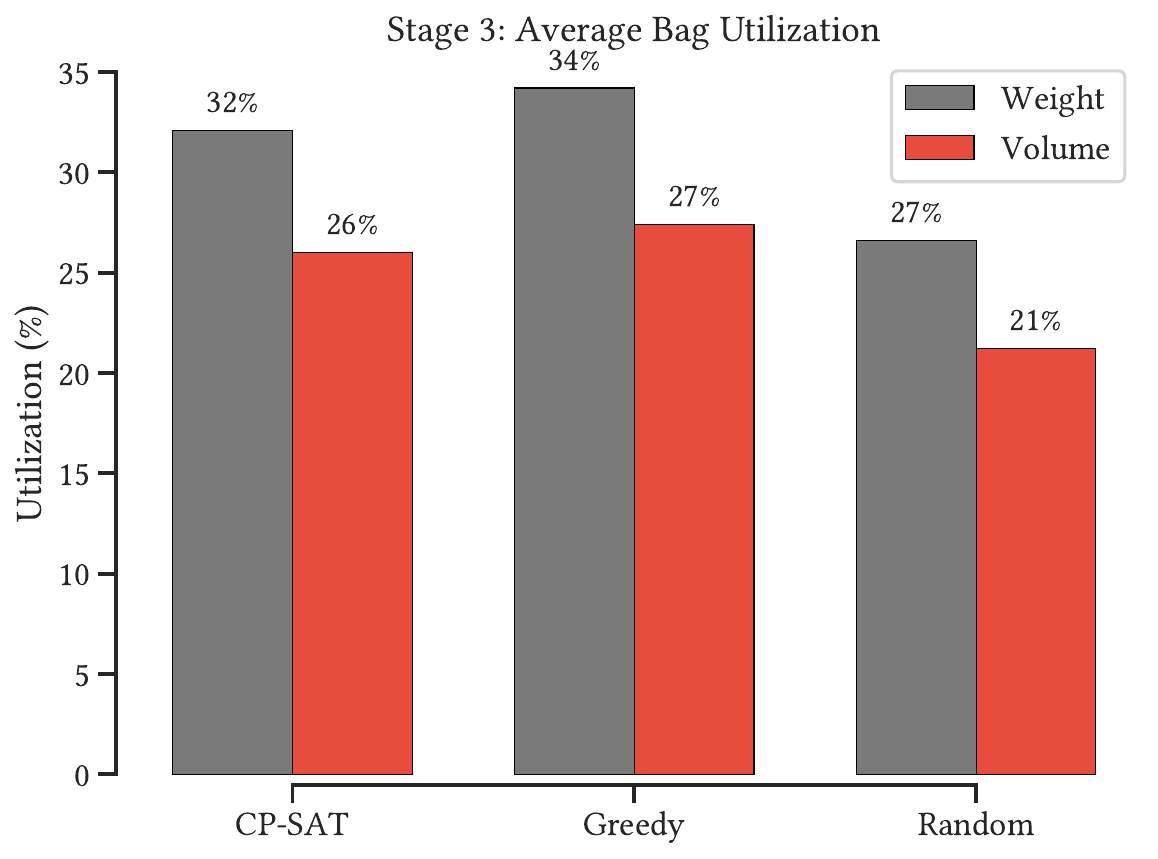}\vspace{-0.05in}
\caption{Average bag utilization by bag type. CP-SAT efficiently uses available capacity while satisfying constraints.\vspace{-0.1in}}
\label{fig:avg_utilization}
\end{minipage}
\end{figure*}

The symbolic engine achieves mean recall of $99.7\%$, with $96.0\%$ of trips having \emph{perfect} recall. Labelers remove $16.7 \pm 22.5$ items per scenario (the large variance reflecting different packing styles, motivating Stage 2 personalization). The moderate $75.7\%$ precision is by design: Stage 2 filters the over-generated seed. Table~\ref{tab:stage1_duration} reports performance by trip duration; recall remains near-perfect across all durations while precision varies from $54.0\%$ (1 to 3 days, where labelers are more selective) to $83.4\%$ (8 to 14 days, where labelers pack more comprehensively).
Acceptance rates vary substantially by item category (Figure~\ref{fig:acceptance_category}): personal items ($78\%$) and clothes ($75\%$) lead; medical ($48\%$) and activity-specific items ($52\%$) trail, reflecting their conditional relevance.


\subsection{Stage 2: Preference Learning}

We compare candidate models for each Stage~2 component using 5-fold trip-level cross-validation, then report combined pipeline recall for the selected configuration.

\noindent\textbf{Inclusion Model Selection.}
Table~\ref{tab:axis1} compares four inclusion models. Gradient boosted trees (GBM, depth=5) achieve the best deployable performance (AUC-ROC $0.943$, Precision $0.969$, Recall $0.813$), outperforming logistic regression and random forests across all metrics. Neural collaborative filtering (NCF) serves as a non-deployable ceiling (AUC-ROC $0.982$): it requires a known user embedding and cannot generalize to cold-start users on mobile.

\begin{figure*}[!htb]
\begin{minipage}[t]{0.31\textwidth}
\centering
\captionsetup{type=table}
\caption{Stage 1 recall, precision, and final checklist size by trip duration. Recall is near-perfect across all durations.\vspace{-0.1in}}
\label{tab:stage1_duration}
\resizebox{\linewidth}{!}{
\begin{tabular}{lccc}
\toprule
\textbf{Duration} & \textbf{Recall} & \textbf{Precision} & \textbf{Final Size} \\
\midrule
1--3 days  & 99.6\% & 54.0\% & $20.3 \pm 15.4$ \\
4--7 days  & 99.6\% & 72.9\% & $30.3 \pm 11.7$ \\
8--14 days & 99.9\% & 83.4\% & $37.4 \pm 8.1$ \\
15+ days   & 100.0\% & 73.2\% & $29.3 \pm 11.3$ \\
\bottomrule
\end{tabular}
}

\end{minipage}
\hspace{.1em}
\begin{minipage}[t]{0.31\textwidth}
\centering
\captionsetup{type=table}
\caption{Inclusion model comparison (mean across 5-fold CV). NCF is only shown as a ceiling (non-deployable).\vspace{-0.1in}}
\label{tab:axis1}
\resizebox{\linewidth}{!}{
\begin{tabular}{lccc}
\toprule
\textbf{Model} & \textbf{AUC-ROC} & \textbf{Precision} & \textbf{Recall} \\
\midrule
Item Frequency         & 0.805 & 0.836 & 0.747 \\
Logistic Regression    & 0.896 & 0.963 & 0.717 \\
Random Forest          & 0.918 & 0.959 & 0.783 \\
\textbf{GBM}           & \textbf{0.943} & \textbf{0.969} & \textbf{0.813} \\
NCF $\dagger$          & 0.982 & 0.976 & 0.928 \\
\bottomrule
\end{tabular}
}
\end{minipage}
\hspace{.1em}
\begin{minipage}[t]{0.31\textwidth}
\centering
\captionsetup{type=table}
\caption{Ranking model comparison (mean across 5-fold CV). LambdaMART achieves the best top-of-list accuracy.\vspace{-0.1in}}
\label{tab:axis2}
\resizebox{\linewidth}{!}{
\begin{tabular}{lccc}
\toprule
\textbf{Model} & \textbf{NDCG@5} & \textbf{NDCG@10} & \textbf{MAP} \\
\midrule
Item Frequency              & 0.908 & 0.892 & 0.860 \\
Ridge Regression            & 0.872 & 0.866 & 0.818 \\
Pairwise LR                 & 0.737 & 0.750 & 0.735 \\
\textbf{LambdaMART}         & \textbf{0.923} & \textbf{0.899} & \textbf{0.864} \\
\bottomrule
\end{tabular}
}
\end{minipage}
\vspace{-0.5em}
\end{figure*}

\noindent\textbf{Ranking Model Selection.}
Table~\ref{tab:axis2} compares four ranking models. LambdaMART achieves the best top-of-list accuracy (NDCG@5 $0.923$, NDCG@10 $0.899$), which matters most since the optimizer uses utility scores to select items under capacity constraints. Item frequency is a strong baseline (NDCG@5 $0.908$), confirming that simple priors capture much of the ranking signal. Pairwise logistic regression performs worst because pairwise binary training does not directly optimize ranking metrics; LambdaMART's NDCG-weighted loss closes this gap.

\begin{figure*}[!htb]
\begin{minipage}{0.31\textwidth}
\centering
\includegraphics[width=\columnwidth]{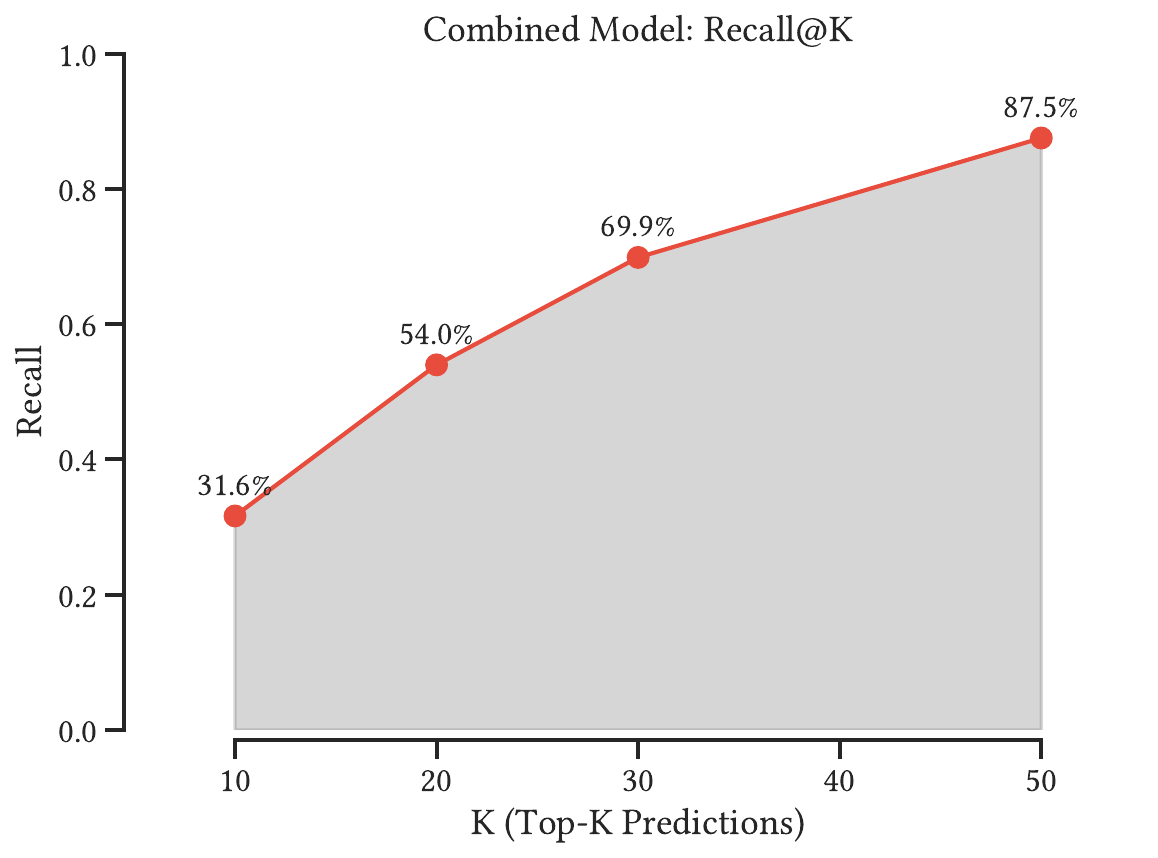}\vspace{-0.05in}
\caption{Combined model Recall@K (5-fold CV). The GBM + LambdaMART pipeline retrieves 87.5\% of top 50 items.\vspace{-0.05in}}
\label{fig:recall_at_k}
\end{minipage}
\hspace{.1em}
\begin{minipage}{0.31\textwidth}
\centering
\includegraphics[width=\columnwidth]{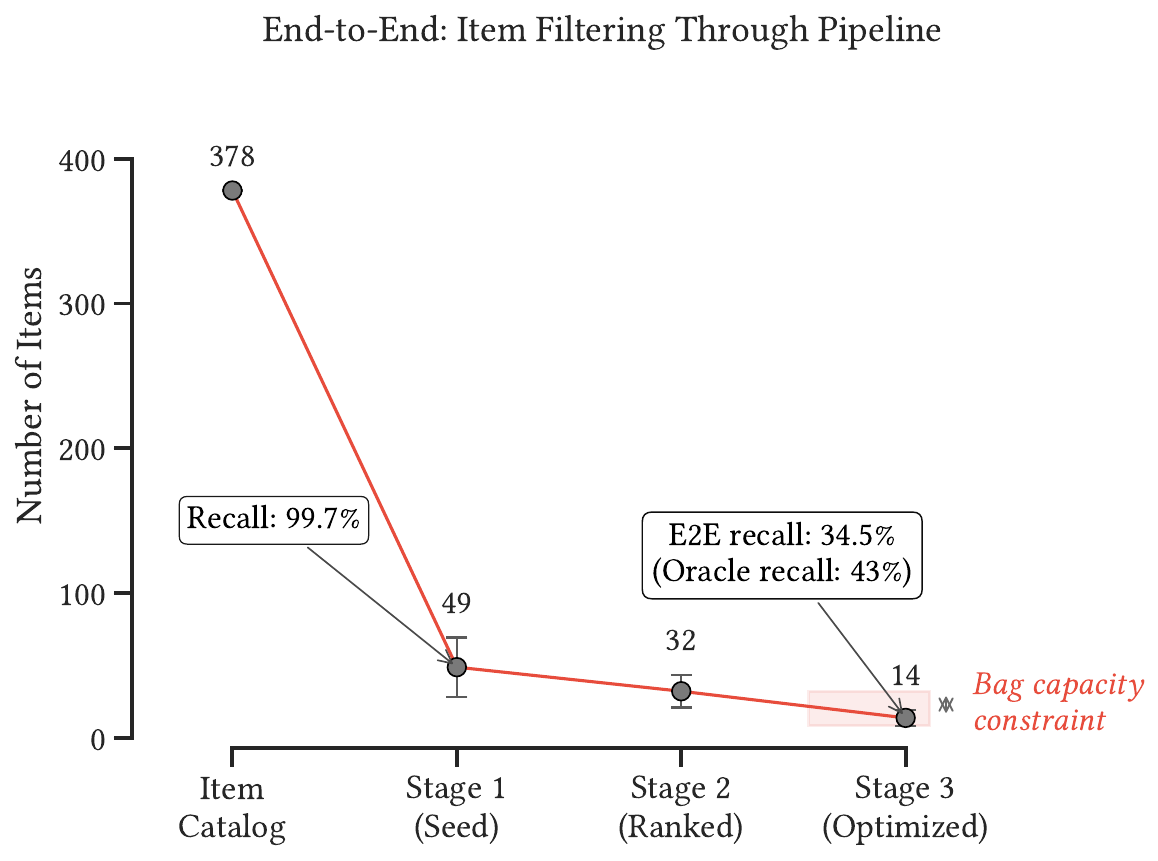}\vspace{-0.05in}
\caption{Item filtering through the pipeline. Each stage progressively refines the item set from catalog to final.\vspace{-0.05in}}
\label{fig:e2e_filtering}
\end{minipage}
\hspace{.1em}
\begin{minipage}{0.31\textwidth}
\centering
\includegraphics[width=\columnwidth]{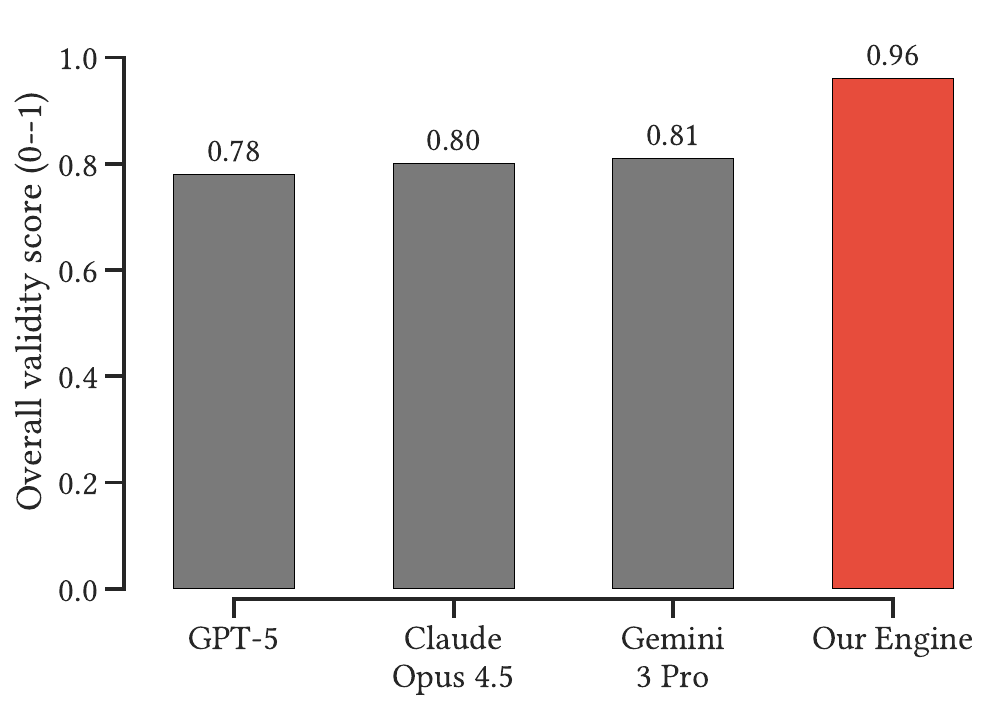}\vspace{-0.05in}
\caption{Overall packing-checklist validity score. Our symbolic engine substantially outperforms frontier LLMs.\vspace{-0.05in}}
\label{fig:packing-overall-valid}
\end{minipage}
\end{figure*}

Figure~\ref{fig:recall_at_k} shows Recall@K for the GBM + LambdaMART pipeline under 5-fold cross-validation. The combined model retrieves $54.0\%$ of labeler-preferred items in the top~20 and $87.5\%$ in the top~50, leaving Stage 3 a high-quality candidate pool to optimize over within its capacity budget. Symbolic provenance features (supporting rules, rule priority, criticality) are the strongest predictors for both models (Figure~\ref{fig:inclusion_features}), empirically confirming that Stage 1 structure carries directly into Stage 2 learning.

\subsection{Stage 3: Constrained Optimization}
The optimization stage assigns items to bags while satisfying hard constraints. We compare CP-SAT against greedy and random baselines; Table~\ref{tab:stage3} reports violations by constraint type. CP-SAT achieves $100\%$ feasibility with zero violations. Greedy achieves only $28\%$ feasibility (883 total violations, mostly must-separate), and Random achieves $10\%$ (1{,}549 violations, on both must-together and must-separate). Neither baseline violates regulatory or capacity constraints, which are checked during single-item assignment; only the dependency constraints, which require coordination across assignments, defeat the heuristics. Figure~\ref{fig:avg_utilization} shows CP-SAT achieves high capacity utilization across all bag types.

We compare four CP-SAT objective formulations to motivate the proposed risk-aware design; all achieve $100\%$ feasibility by construction. Variants: (\textit{i})~\textbf{EV} (proposed): maximize expected retained utility minus expected monetary loss; (\textit{ii})~\textbf{MinimaxRegret}: minimize worst-case regret over bag-loss scenarios; (\textit{iii})~\textbf{Survival}: EV with hard constraints banning critical items from high-loss bags; (\textit{iv})~\textbf{Sequential}: greedy risk-corrected selection followed by CP-SAT bag assignment. Table~\ref{tab:formulations} reports outcomes.
EV achieves the highest retained utility and is our proposed formulation. Sequential yields essentially identical outcomes, suggesting the simpler EV objective is sufficient. MinimaxRegret is worse on all three reported metrics (utility, tail loss, critical-item survival) and takes nearly twice as long to solve. Survival sacrifices roughly $13\%$ retained utility (220.0 vs. EV's 253.6) for the lowest tail loss and highest critical-item survival, an appropriate trade when critical-item protection is paramount.

\begin{figure*}[!htb]
\begin{minipage}[t]{0.31\textwidth}
\centering
\captionsetup{type=table}
\caption{Constraint violations for optimization methods, by constraint type. Only CP-SAT achieves zero violations.\vspace{-0.05in}}
\label{tab:stage3}
\resizebox{\linewidth}{!}{
\begin{tabular}{lccc}
\toprule
\textbf{Optimization} & \textbf{Packing} & \textbf{Must} & \textbf{Must} \\
\textbf{Method} & \textbf{Feasibility} & \textbf{Together} & \textbf{Separate} \\
\midrule
CP-SAT  & 100\% & 0     & 0   \\
Greedy  & 28\%  & 79    & 804 \\
Random  & 10\%  & 1,038 & 511 \\
\bottomrule
\end{tabular}
}
\end{minipage}
\hspace{.1em}
\begin{minipage}[t]{0.31\textwidth}
\centering
\captionsetup{type=table}
\caption{CP-SAT optimization formulation comparison.  EV achieves the highest retained utility.\vspace{-0.1in}}
\label{tab:formulations}
\resizebox{\linewidth}{!}{
\begin{tabular}{lccc}
\toprule
\textbf{Formulation} & \textbf{Retained} & \textbf{CVaR-90} & \textbf{P(Critical} \\
                      & \textbf{Utility}    & \textbf{Loss}    & \textbf{Survived)} \\
\midrule
EV (proposed)      & \textbf{253.6} & 17.0 & 98.6\% \\
MinimaxRegret      & 252.8 & 24.7 & 98.0\% \\
Survival           & 220.0 & \textbf{11.9} & \textbf{99.5\%} \\
Sequential         & 253.5 & 17.0 & 98.6\% \\
\bottomrule
\end{tabular}
}
\end{minipage}
\hspace{.1em}
\begin{minipage}[t]{0.31\textwidth}
\centering
\captionsetup{type=table}
\caption{Post-deployment engagement changes in \textsc{FlyEnJoy} relative to the prior template-based baseline.\vspace{-0.1in}}
\label{tab:post_deployment_packing}
\resizebox{\linewidth}{!}{
\begin{tabular}{llc}
\toprule
\textbf{Metric} & \textbf{Type} & \textbf{Change} \\
\midrule
Checklist Initiations     & Usage      & $+200\%$  \\
Completed Checklists      & Usage      & $+100\%$  \\
DAU/WAU Ratio             & Engagement & $+29$ pp  \\
Time per Session          & Efficiency & $-49.6\%$ \\
\bottomrule
\end{tabular}
}

\end{minipage}
\vspace{-0.5em}
\end{figure*}

\subsection{End-to-End Evaluation}
We evaluate the complete pipeline with 5-fold cross-validation on 604 scenarios; Stage 2 is trained on training trips and the full pipeline is evaluated on held-out trips. Figure~\ref{fig:e2e_filtering} traces an item through the pipeline: 378 catalog items, 49 in the Stage 1 seed, 32 in the labeler reference, and 14 selected by Stage 3 under capacity and dependency constraints.
Because the optimizer packs $\approx 14$ items against a $\approx 32$-item labeler reference, no system, however accurate, can exceed an oracle recall of $14/32 \approx 43\%$. The pipeline achieves precision $80.9\% \pm 25.5\%$ and recall $34.5\% \pm 14.3\%$, recovering roughly $80\%$ of the oracle maximum. Recall in this setting is bounded by capacity, not by model quality.

We compare our symbolic engine against frontier LLMs using a four-component validity rubric: (i) TSA compliance, (ii) weather appropriateness, (iii) activity requirements, and (iv) completeness, each scored in $[0,1]$ with overall score being their average. Our engine achieves $0.96$, substantially outperforming Gemini 3 Pro ($0.81$), Claude Opus 4.5 ($0.80$), and GPT-5 ($0.78$) (Figure~\ref{fig:packing-overall-valid}). LLM outputs average 36 to 45 items per scenario, missing rubric-required items that our engine includes; our engine deliberately over-generates (69 to 81 items) to feed Stage 2 filtering.

\subsection{Post-Deployment Quantification}
We deployed the symbolic reasoning and preference learning stages within FlyEnJoy, a production iOS travel application, and evaluated post-launch impact using anonymized interaction logs. The prior deployment was a static, template-based checklist with no personalization or constraint optimization; the current system replaces the underlying logic but preserves the user interface, so observed changes reflect personalization quality rather than UI redesign. Due to business sensitivity, we report relative changes rather than absolute counts.
Feature engagement increased substantially (Table~\ref{tab:post_deployment_packing}). Checklist initiations rose by $200\%$ and completed checklists by $100\%$, indicating both improved discovery and task follow-through. Per-session time decreased sharply, driven by a sharp drop in manual edits: users now reach a satisfactory checklist with substantially fewer add and remove actions per session, our direct proxy for personalization quality. Together, the adoption gains and the reduction in editing effort confirm that the framework delivers measurable real-world value over the prior template-only baseline.



\section{Related Work}
Existing works on packing checklist recommendations offer little personalization and do not explicitly model hard constraints. Learning-based approaches infer item co-occurrence from past lists~\cite{ravi2025multi} but produce unconstrained rankings. Travel-planning assistants including SwiftTrip~\cite{regin2024swifttrip}, TripEase GenAI~\cite{lim2024enhancing}, and Tripmate~\cite{nikam2025interactive} generate packing lists from destination context and weather, but treat packing as an auxiliary feature with weak feasibility guarantees. Clothing-oriented systems focus on outfit recommendation rather than checklist feasibility~\cite{zhang2017trip}; luggage packing algorithms optimize geometric placement of fixed item sets~\cite{tiwari2010fast}. In contrast, we generate personalized checklists that are feasible by construction under safety, capacity, and dependency constraints.

Methodologically, our framework reconciles three research lines that have not previously been combined for personalized constrained subset selection. Implicit-feedback recommenders learn preferences from interaction traces~\cite{hu2008collaborative,rendle2009bpr,he2017ncf,liang2018variational}; learning-to-rank methods such as LambdaMART optimize ranking metrics directly~\cite{liu2009learning,burges2010ranknet}, but both assume an unconstrained candidate space. Knowledge- and constraint-based recommenders encode domain knowledge via rules or constraint satisfaction, improving interpretability and cold-start robustness, but typically require manually elicited preferences rather than learning them from behavior~\cite{aggarwal2016knowledge,uta2024knowledge,felfernig2015constraint}. LLM-based recommenders provide text-grounded explanations but cannot reliably enforce hard constraints and are unsuitable for offline, on-device deployment~\cite{zhao2024recommender}. We combine the strengths of each line through neural-symbolic decomposition with CP-SAT optimization~\cite{garcez2019neural,rossi2006handbook}.

\section{Conclusion}
\label{sec:conclusion}
We presented a three-stage framework that combines symbolic rule-based candidate generation, survivorship-bias-aware preference learning, and CP-SAT constrained optimization to produce personalized packing checklists that are feasible by construction. On 604 trip scenarios, the framework attains 99.7\% candidate recall, AUC-ROC 0.943 for inclusion, NDCG@5 0.923 for ranking, and 100\% constraint satisfaction. Deployment in FlyEnJoy, a production iOS travel app, doubled checklist completions and reduced editing and completion time.
The end-to-end pipeline achieves 80.9\% precision against a labeler reference of $\approx$32 items while packing only $\approx$14 items under capacity, recovering roughly 80\% of the oracle-feasible maximum.
The architecture instantiates a general pattern for constrained personalization: encode hard rules symbolically, learn preferences from behavior, and reconcile them through constrained optimization. The pattern transfers naturally to settings where compliance and personalization coexist, such as clinical discharge planning and immigration documentation. The principal limitation is the cost of authoring and maintaining domain rules; we view rule reuse across families as a partial mitigation and a direction for future work.

\section*{GenAI Usage Disclosure}
During the preparation of this work, the authors used LLM-based tools to assist with condensing and improving the clarity of the author-written text within selected sections. No new content, results, analyses or claims were created by AI tools. All AI-assisted edits were reviewed, verified and approved by the authors. The authors take full responsibility for the accuracy and integrity of all content in this paper.

\balance

\bibliographystyle{ACM-Reference-Format}
\bibliography{references,checklist}

@article{ravi2025multi,
  title={A Multi-tiered Solution for Personalized Baggage Item Recommendations using FastText and Association Rule Mining},
  author={Ravi, Mudavath and Negi, Atul},
  journal={arXiv preprint arXiv:2501.09359},
  year={2025}
}

@article{uta2024knowledge,
  title={Knowledge-based recommender systems: overview and research directions},
  author={Uta, Mathias and Felfernig, Alexander and Le, Viet-Man and Tran, Thi Ngoc Trang and Garber, Damian and Lubos, Sebastian and Burgstaller, Tamim},
  journal={Frontiers in big Data},
  volume={7},
  pages={1304439},
  year={2024},
  publisher={Frontiers Media SA}
}

@incollection{felfernig2015constraint,
  title={Constraint-based recommender systems},
  author={Felfernig, Alexander and Friedrich, Gerhard and Jannach, Dietmar and Zanker, Markus},
  booktitle={Recommender systems handbook},
  pages={161--190},
  year={2015},
  publisher={Springer}
}

@article{zhao2024recommender,
  title={Recommender systems in the era of large language models (llms)},
  author={Zhao, Zihuai and Fan, Wenqi and Li, Jiatong and Liu, Yunqing and Mei, Xiaowei and Wang, Yiqi and Wen, Zhen and Wang, Fei and Zhao, Xiangyu and Tang, Jiliang and others},
  journal={IEEE Transactions on Knowledge and Data Engineering},
  volume={36},
  number={11},
  pages={6889--6907},
  year={2024},
  publisher={IEEE}
}

@article{regin2024swifttrip,
  title={SwiftTrip: Your Smart Travel Companion for Effortless Planning and Memorable Journeys},
  author={Regin, R and Rajest, S Suman},
  journal={International Journal of Human Computing Studies},
  volume={6},
  number={3},
  pages={57--70},
  year={2024}
}

@article{lim2024enhancing,
  title={Enhancing Travel Planning Efficiency with a Comprehensive TripEase GenAI Mechanism.},
  author={Lim, Zhi-Sheng and Yulastri, Asmar and Ho, Sin-Ban and Tan, Chuie-Hong},
  journal={International Journal on Advanced Science, Engineering \& Information Technology},
  volume={14},
  number={6},
  year={2024}
}

@article{nikam2025interactive,
  title={An Interactive Mobile Application to Enhance Travel Planning Using a User-Centric Approach: A UX Design Case Study.},
  author={Nikam, Avishkar and Verma, Indresh Kumar and Ranade, Pranita},
  journal={International Journal of Interactive Mobile Technologies},
  volume={19},
  number={17},
  year={2025}
}

@article{zhang2017trip,
  title={Trip outfits advisor: Location-oriented clothing recommendation},
  author={Zhang, Xishan and Jia, Jia and Gao, Ke and Zhang, Yongdong and Zhang, Dongming and Li, Jintao and Tian, Qi},
  journal={IEEE Transactions on Multimedia},
  volume={19},
  number={11},
  pages={2533--2544},
  year={2017},
  publisher={IEEE}
}

@article{tiwari2010fast,
author = {Tiwari, Santosh and Fadel, Georges and Fenyes, Peter},
year = {2010},
month = {06},
pages = {},
title = {A Fast and Efficient Compact Packing Algorithm for SAE and ISO Luggage Packing Problems},
volume = {10},
journal = {Journal of Computing and Information Science in Engineering - JCISE},
doi = {10.1115/1.3330440}
}

@misc{tsa2026whatcanibring,
  author       = {{Transportation Security Administration}},
  title        = {What Can I Bring?},
  howpublished = {\url{https://www.tsa.gov/travel/security-screening/whatcanibring/all}},
  year         = {2026},
  note         = {Accessed: 2026-02-08},
  publisher    = {U.S. Department of Homeland Security}
}

@misc{tsa2026advisories,
  author       = {{Transportation Security Administration}},
  title        = {Travel Advisories},
  howpublished = {\url{https://travel.state.gov/en/international-travel/travel-advisories.html}},
  year         = {2026},
  note         = {Accessed: 2026-02-08},
  publisher    = {U.S. Department of Homeland Security}
}

@misc{faa2026lithium,
  author       = {{Federal Aviation Administration}},
  title        = {Lithium Batteries},
  howpublished = {\url{https://www.faa.gov/hazmat/packsafe/lithium-batteries}},
  year         = {2026},
  note         = {Accessed: 2026-02-08},
  publisher    = {U.S. Department of Transportation}
}

@misc{christine2024ultimate,
  author       = {{Sarkis, Christine}},
  title        = {The Ultimate Packing List},
  howpublished = {\url{https://www.smartertravel.com/the-ultimate-packing-list/}},
  year         = {2024},
  note         = {Accessed: 2026-02-08}
}

@misc{Canva2026free,
  author       = {{Canva}},
  title        = {Free and Customizable Travel Checklist Templates},
  howpublished = {\url{https://www.canva.com/checklists/templates/travel/}},
  year         = {2026},
  note         = {Accessed: 2026-02-08}
}

@misc{talkerresearch2024forget,
  title = {Survey: Nearly 90\% of Americans Forget Essential Items While Traveling},
  author = {{Talker Research}},
  year = {2024},
  howpublished = {\url{https://www.talkerresearch.com/}},
  note = {Survey of 2,000 U.S. adults}
}

@inproceedings{burges2010ranknet,
  title={From RankNet to LambdaRank to LambdaMART: An overview},
  author={Burges, Christopher JC},
  booktitle={Learning},
  volume={11},
  number={23-581},
  pages={81},
  year={2010},
  publisher={Microsoft Research Technical Report}
}

@article{liu2009learning,
  title={Learning to rank for information retrieval},
  author={Liu, Tie-Yan},
  journal={Foundations and Trends in Information Retrieval},
  volume={3},
  number={3},
  pages={225--331},
  year={2009},
  publisher={Now Publishers, Inc.}
}

@inproceedings{rendle2009bpr,
  title={{BPR}: Bayesian personalized ranking from implicit feedback},
  author={Rendle, Steffen and Freudenthaler, Christoph and Gantner, Zeno and Schmidt-Thieme, Lars},
  booktitle={Proceedings of the Twenty-Fifth Conference on Uncertainty in Artificial Intelligence},
  pages={452--461},
  year={2009}
}

@inproceedings{he2017ncf,
  title={Neural collaborative filtering},
  author={He, Xiangnan and Liao, Lizi and Zhang, Hanwang and Nie, Liqiang and Hu, Xia and Chua, Tat-Seng},
  booktitle={Proceedings of the 26th International Conference on World Wide Web},
  pages={173--182},
  year={2017}
}

@inproceedings{hu2008collaborative,
  title={Collaborative filtering for implicit feedback datasets},
  author={Hu, Yifan and Koren, Yehuda and Volinsky, Chris},
  booktitle={2008 Eighth IEEE International Conference on Data Mining},
  pages={263--272},
  year={2008},
  organization={IEEE}
}

@inproceedings{aggarwal2016knowledge,
  title={Knowledge-based recommender systems},
  author={Aggarwal, Charu C},
  booktitle={Recommender Systems},
  pages={167--197},
  year={2016},
  publisher={Springer}
}

@article{garcez2019neural,
  title={Neural-Symbolic Computing: An Effective Methodology for Principled Integration of Machine Learning and Reasoning},
  author={Artur S. d'Avila Garcez and Marco Gori and L. Lamb and Luciano Serafini and Michael Spranger and S. Tran},
  journal={FLAP},
  year={2019},
  volume={6},
  pages={611-632},
  url={https://api.semanticscholar.org/CorpusID:155092677}
}

@article{rossi2006handbook,
  title={Handbook of constraint programming},
  author={Rossi, Francesca and Van Beek, Peter and Walsh, Toby},
  year={2006},
  publisher={Elsevier}
}

@inproceedings{liang2018variational,
  title={Variational autoencoders for collaborative filtering},
  author={Liang, Dawen and Krishnan, Rahul G and Hoffman, Matthew D and Jebara, Tony},
  booktitle={Proceedings of the 2018 World Wide Web Conference},
  pages={689--698},
  year={2018}
}



\end{document}